%% file: root_journal.tex
\documentclass[lettersize,journal]{IEEEtran}
\usepackage{amsmath,amsfonts}
\usepackage{array}
\usepackage{stfloats}
\usepackage{url}
\usepackage{verbatim}
\usepackage{graphicx}
\usepackage{cite}
\usepackage{mathptmx} %
\usepackage{amssymb}  %

\usepackage{paralist}
\usepackage{color}
\usepackage{multirow}
\usepackage{hyperref}
\usepackage{subcaption}
\usepackage{algorithm,algpseudocode}
\usepackage{xr}

\usepackage{siunitx}
\usepackage{mathtools}
\usepackage{wrapfig}
\usepackage{newtxtext,newtxmath}
\usepackage{comment}
\usepackage{ifthen}
\usepackage{cleveref}

\usepackage{booktabs}
\usepackage{lipsum}  
\usepackage[export]{adjustbox}
\usepackage{threeparttable}
\usepackage{adjustbox}

\crefformat{section}{\S#2#1#3} %
\crefformat{subsection}{\S#2#1#3}
\crefformat{subsubsection}{\S#2#1#3}

\usepackage{epstopdf}

\DeclareGraphicsExtensions{.pdf, .png}

\newcommand{\JournalFormat}{true}
\newcommand{\IfJournal}[1]{
    \ifthenelse{\equal{\JournalFormat}{true}}{#1}{}
}
\newcommand{\IfJournalThenElse}[2]{
    \ifthenelse{\equal{\JournalFormat}{true}}{#1}{#2}
}
\newcommand{\IfConference}[1]{
    \ifthenelse{\equal{\JournalFormat}{false}}{#1}{}
}

\newcommand{\ShowComments}{false} 
\newcommand{\SR}[1]{
    \ifthenelse{\equal{\ShowComments}{true}}{\noindent\textcolor{red}{Sadegh: #1}}{}
}
\newcommand{\JB}[1]{
    \ifthenelse{\equal{\ShowComments}{true}}{\noindent\textcolor{blue}{Joydeep: #1}}{}
}
\newcommand{\CB}[1]{
    \ifthenelse{\equal{\ShowComments}{true}}{\noindent\textcolor{green}{Connor: #1}}{}
}
\newcommand{\KV}[1]{
    \ifthenelse{\equal{\ShowComments}{true}}{\noindent\textcolor{pink}{Kavan: #1}}{}
}

\newcommand{\changesColor}{black}
\newcommand{\HighlightRevisions}{false} 
\newcommand{\changes}[1]{
    \ifthenelse{\equal{\HighlightRevisions}{true}}{\textcolor{blue}{#1}}{\textcolor{black}{#1}}
}
\newcommand{\changesCaption}[1]{
    \ifthenelse{\equal{\HighlightRevisions}{true}}{\caption{\textcolor{blue}{#1}}}{\noindent\caption{\textcolor{black}{#1}}}
}

\newcommand{\thisWork}{CPIP}

\newcolumntype{M}[1]{>{\centering\arraybackslash}m{#1}}

\DeclareMathOperator*{\bel}{bel}

\algnewcommand\algorithmicforeach{\textbf{for each}}
\algdef{S}[FOR]{ForEach}[1]{\algorithmicforeach\ #1\ \algorithmicdo}

\newcommand*{\compModel}{h}

\newcommand*{\f}{f_{i, e}}
\newcommand{\belOf}[2]{{\bel}_{#1}(#2)}

\begin{document}

\title{
Competence-Aware Path Planning via \\ Introspective Perception
}

\author{Sadegh Rabiee$^{1}$ \and Connor Basich$^{2}$ \and Kyle Hollins Wray$^{3}$ \and Shlomo Zilberstein$^{2}$ \and Joydeep Biswas$^{1}$ %
\thanks{Manuscript received: September 9, 2021; Revised December, 9, 2021; Accepted January 4, 2022.}%
\thanks{This paper was recommended for publication by Editor Editor Cesar Cadena upon evaluation of the Associate Editor and Reviewers' comments.
This work was supported in part by NSF (CAREER-2046955,
IIS-1954778) and DARPA (HR001120C0031). The views and conclusions contained in this document are those of the authors only.} %
\thanks{$^{1}$Sadegh Rabiee and Joydeep Biswas are with the Department of Computer Science, The University of Texas at Austin, Austin, TX. Email:
        {\tt\small \{srabiee, joydeepb\}@cs.utexas.edu }}
\thanks{$^{2}$Connor Basich and Shlomo Zilberstein are with the College of Information and Computer Sciences, The University of Massachusetts Amherst, Amherst, MA. Email:
        {\tt\small \{cbasich, shlomo\}@cs.umass.edu }}
\thanks{$^{3}$Kyle Hollins Wray is with the Alliance Innovation Lab Silicon Valley, Santa Clara, CA. Email:
        {\tt\small  kyle.wray@nissan-usa.com}}
\thanks{Digital Object Identifier (DOI): see top of this page.}
}

\markboth{IEEE Robotics and Automation Letters. Preprint Version. Accepted January, 2022}
{Rabiee \MakeLowercase{\textit{et al.}}: Competence-Aware Path Planning}

\maketitle

\begin{abstract}
\input{abstract}

\end{abstract}

\input{introduction}

\input{related_work}

\input{method}

\input{experimental_results}

\input{conclusion}

\bibliographystyle{IEEEtran}
\bibliography{IEEEabrv,bibliography}

\end{document}

%% file: abstract.tex
Robots deployed in the real world over extended periods of time need to reason about unexpected failures, learn to predict them, and to proactively take actions to avoid future failures.
Existing approaches for competence-aware planning are either model-based, requiring explicit enumeration of known failure sources, or purely statistical,  using state- and location-specific failure statistics to infer competence.
We instead propose a structured model-free approach to competence-aware planning by reasoning about
plan execution failures due to errors in perception, without requiring a priori enumeration of failure sources or requiring location-specific failure statistics.
We introduce \emph{competence-aware path planning via introspective perception (CPIP)}, a Bayesian framework to iteratively learn and exploit task-level competence in novel deployment environments.
CPIP factorizes the competence-aware planning problem into two components.
First, perception errors are learned in a model-free and location-agnostic setting via \emph{introspective perception} prior to deployment in novel environments.
Second, during actual deployments, the prediction of task-level failures is learned in a context-aware setting.
Experiments in a simulation show that the proposed CPIP approach outperforms the frequentist baseline in multiple mobile robot tasks, and is further validated via real robot experiments in environments with perceptually challenging obstacles and terrain.

%% file: introduction.tex
\section{Introduction}
\IfJournalThenElse{\IEEEPARstart{A}{s}}{As}robots become increasingly available, they are deployed for tasks where autonomous navigation in uncontrolled environments is crucial to success, such as package delivery, warehouse automation, and home service settings.
Deploying robots over extended periods of time and in such open world setting requires addressing failures originating from real-world uncertainty and imperfect perception. 
Continuous operator monitoring, while effective, is cumbersome and thus not scalable to many robots or large environments. 
We are thus interested in developing \emph{competence-aware} agents capable of assessing the probability of successfully completing a given task. Such agents would learn from failures and leverage the acquired knowledge when planning to improve their robustness and reliability.  
Previous efforts towards competence-aware path planning and motion planning either rely solely on statistical analysis of logged instances of failures in the configuration space of the robot and do not benefit from the sensing information collected by the robot~\cite{hawes2017strands}, or are application specific and designed to reduce the probability of failure for a specific perception module such as visual SLAM~\cite{costante2016perception}.
While there has been progress on introspective perception to enable perception algorithms to learn to predict their sources of errors~\cite{rabiee2019ivoa, rabiee2020ivslam}, the outputs of such algorithms have not yet been exploited in robot planning.

We present competence-aware path planning via introspective perception (\thisWork), a general framework that bridges the gap between path planning and introspective perception and allows the robot to iteratively learn and exploit task-level competence in novel deployment environments.
\thisWork{} models the path planning problem as a Stochastic Shortest Path (SSP) problem and builds a model that represents both the topological map of the environment as well as the competence of the robot in traversing each part of the map autonomously. 
\thisWork{} leverages introspective perception to predict the task-level competence of the robot in novel deployment environments and employs a Bayesian approach to update its estimate of the robot competence online and during the deployment. \thisWork~ then uses this information to plan paths that reduce the risk of failures.

Our experimental results demonstrate that \thisWork~ converges to the optimal planning policy in novel deployment environments while reducing the frequency of navigation failures by more than $80\%$ compared to the state-of-the-art competence-aware path planning algorithms that do not leverage introspective perception.

%% file: related_work.tex
\section{Related Work} \label{related_work}

The idea of integrating perception with planning and control was introduced by pioneering works on active perception that suggested performance of perception can be improved by selecting control strategies that depend on the current state of perception data interpretation as well as the goal of the task~\cite{bajcsy1988active, aloimonos1988active}. Researchers have applied this idea to various levels of control ranging from active vergence control for a stereo pair of cameras~\cite{krotkov1988focusing} to
object manipulation given the next best view for surface reconstruction of unknown objects~\cite{krainin2011autonomous}. 

One line of work predicts and avoids degradation of perception performance given features extracted from the raw sensory data.
Costante et al.~\cite{costante2016perception} propose a perception-aware path planning for MAVs that maximizes the information gain from image matching while solving for dense V-SLAM. Sadat et al.~\cite{sadat2014feature} and Deng et al.~\cite{deng2018feature} follow a similar approach and use an RRT* planner where the cost of a path is defined as a linear combination of the length of the path and the predicted density of image features along the path to reduce localization errors. 
In these works, estimates of competence for perception are obtained via hand crafted metrics and the path planner cost function is designed to specifically address the reliability of V-SLAM and is not generalizable to arbitrary perception tasks. 

\changes{There exists a body of work on risk-aware path planning, where it is assumed that the autonomous agent has accurate models of the uncertainty of perception. Jasour et al.~\cite{jasour2019risk} use the \emph{a priori} known parametric probability distributions for obstacle locations and leverage chance constrained optimization to plan paths that have collision probabilities below a user-specified threshold. Schirmer et al.~\cite{shah2018airsim} use an offline-built localization uncertainty map of the environment to do risk-aware path planning.
Barbosa et al.~\cite{barbosa2021risk} and similarly Chung et al.~\cite{chung2019risk} relax the full-observability assumption and do path planning in partially known environments, yet they assume that the agent has an observation model with a known noise process that is used to update its belief over the state of the world. However, in practice estimates of uncertainty of perception algorithms, as obtained by methods such as the Cramer-Rao lower bound, are often overconfident and inaccurate. An alternative approach is to directly model task-level failures as a function of the state of the world as acquired by perception.
Saxena et al.~\cite{saxena2017learning} learn to predict task-level failures that are due to errors in perception from the raw sensory data; however, predicted failures are used to trigger an enumerated set of recovery actions rather than proactively generating plans that reduce the probability of failures. Similarly Gurau et al.~\cite{guruau2018learn} leverage image data and location specific features to do reactive planning by selecting between different levels of autonomy at any point in time.}

A different line of work on competence-aware path planning that has a more holistic view of failures includes keeping track of all of the robot failures regardless of the perception algorithm that is the cause, and then leveraging this information to proactively generate plans with reduced risk of failures.
Lacerda et al.~\cite{lacerda2014optimal} aggregate the failure instances of a service mobile robot while navigating the environment to model the probability of success for traversing each edge of a topological map using an MDP and generate navigation policies that prefer paths with high success probabilities.  Krajník~\cite{krajnik2017fremen} use a spectral model to learn mid to long-term environmental changes assuming they have a periodic nature and exploit it to improve robot navigation and localization by predicting such changes. Vintr et al.~\cite{vintr2019spatio} use a similar approach to learn a spatio-temporal model for predicting presence of humans in the robot's deployment environment at different times of the day. 
Since these methods are based on statistical analysis of the frequency of navigation failures, they require ample experience and several samplings from any location in the map, in order to achieve an accurate estimate of the robot's competence in navigating that specific location. Moreover, due to using location specific features of the environment for estimating the robot competence, these estimates cannot be generalized to novel deployment environments.
Basich et al.~\cite{basich2020learning} further expand the concept of competence to the optimal level of autonomy and define a stochastic model for solving the path planning problem, where the generated plans consist of a path and the optimal level of autonomy for each segment of the path. In order to learn to predict the probability of failure at each level of autonomy this work requires a curated list of environmental features that are potentially correlated with robot failures.

In this work we leverage machine learned models capable of predicting errors of individual perception modules, to obtain an accurate estimate of the robot's competence at successfully navigating throughout an environment. \thisWork{} uses the estimate of competence to plan reliable and short-duration paths. 
Our work is similar to~\cite{lacerda2014optimal, basich2020learning} in that it reasons about the competence of the robot at successfully performing navigation tasks at a topological map level;
however, it removes the need for an enumerated list of perception related features by automatically learning to extract such features from the raw sensory data. 
Furthermore, \thisWork{} significantly reduces the frequency of failures experienced in new environments by exploiting the generalizable learned perception features instead of merely relying on statistical analysis of the location of previous navigation failures.

%% file: method.tex
\section{CPIP Definition} \label{sec:cpip_definition}

\thisWork{} is a framework for integrating path planning with introspective perception in life-long learning settings. It is defined as a tuple $<\mathcal{M}, \mathcal{I}, \mathcal{H}>$, where $\mathcal{M}$ is a stochastic planning model, $\mathcal{I} = \{\mathbb{I}_k\}_{k=1}^{N}$ is a set of introspective perception modules, and $\mathcal{H}$ is a task-level competence predictor.
\thisWork{} leverages introspective perception and the competence predictor model to predict the probability of task-level failures given the raw sensory data at every time step and uses these estimates to update the planning model iteratively during robot deployments, hence learns policies that reduce the probability of failures. 
In~\cref{sec:planning}, we introduce the planning model $\mathcal{M}$ and explain how it incorporates the probability of autonomous navigation failure in path planning. We then explain introspective perception $\mathcal{I}$ and the competence predictor model $\mathcal{H}$ and how they are used to structure the problem of learning to predict instances of navigation failures in~\cref{sec:introspection}.

\section{Competence-Aware Planning} \label{sec:planning}
The \thisWork{} planning model $\mathcal{M}$ uses a representation of the environment that includes both the connectivity of a set of sparse locations on the map as well as the probability of successful traversal between each two connected neighboring locations.
In this section, we explain this model and how it is actively updated during deployments.

\subsection{Planning Model Description} \label{sec:planning_background}
\begin{figure}[t]
\centering
\includegraphics[width=0.95\linewidth, trim=100 130 200 0,clip]{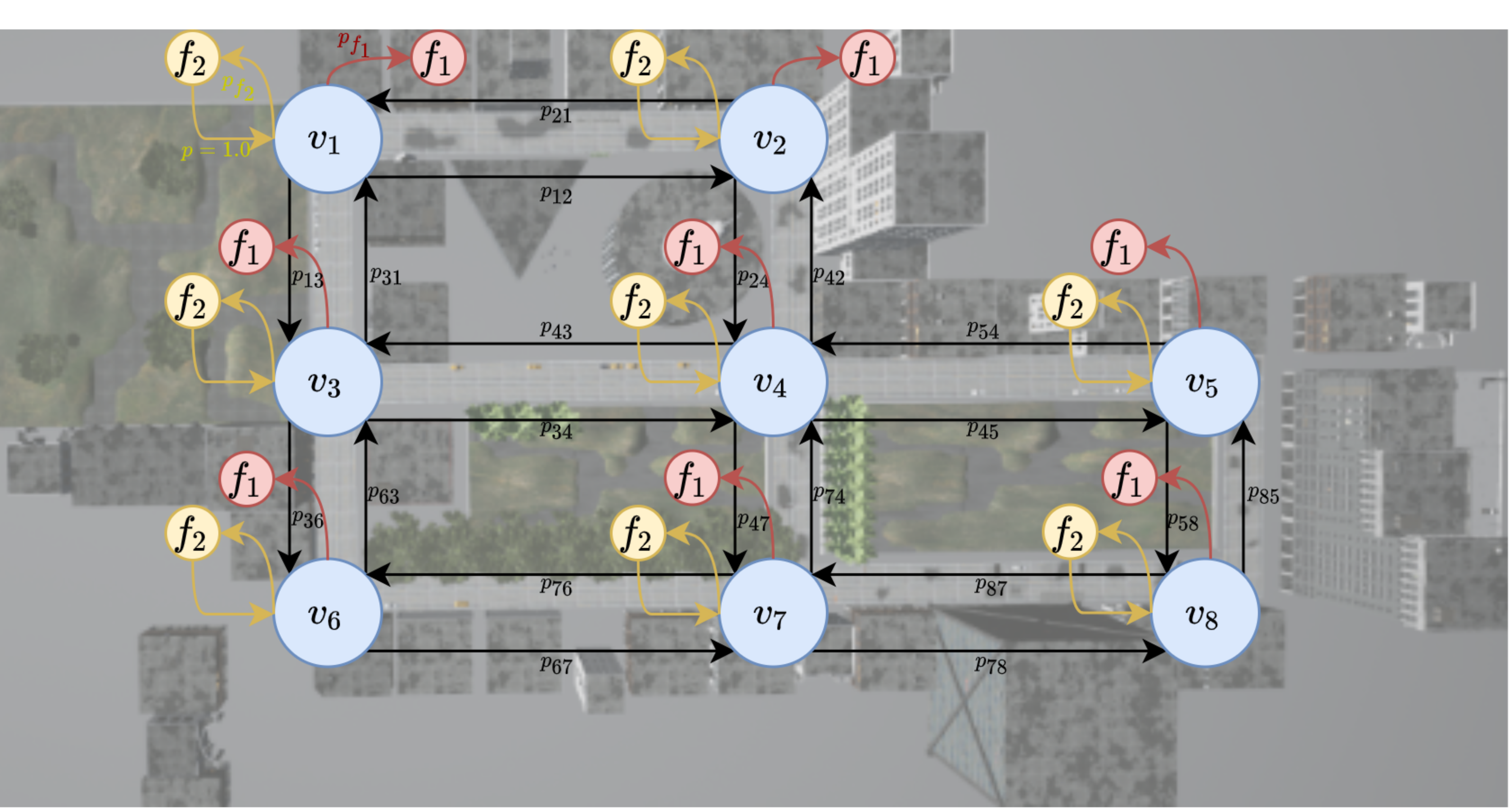}
\caption{Planning SSP for an example environment.}\label{fig:cpip_mdp_diagram}
\IfConference{\vspace{-1em}}
\end{figure}

The input to our problem is a topological map of the environment in the form of a directed graph $G=\langle N,E\rangle$ composed of a set of nodes, $N$, and a set of edges, $E$. Each node represents a location, and each edge $e$ is defined by a tuple $\langle n_e^i, n_e^j, t_e, p_e \rangle$. Here, $n_e^i$ is the starting vertex, $n_e^j$ is the ending vertex, $t_e$ is the expected traversal time for the edge $e$, and $p_e$ is the probability of successfully traversing the edge.

Given the topological map, $G$, we model the planning problem as a Stochastic Shortest Path (SSP) problem, a formal decision-making model for reasoning in stochastic environments where the objective is to find the least-cost path from a start state to a goal state. An SSP is a tuple $\langle S, A, T, C, s_0, G \rangle$ where $S$ is a finite set of states, $A$ is a finite set of actions, $T : S \times A \times S \rightarrow [0,1]$ represents the probability of reaching state $s' \in S$ after performing action $a \in A$ in state $s \in S$, $C : S \times A \rightarrow \mathbb{R}^+$ represents the expected immediate cost of performing action $a \in A$ in state $s \in S$, $s_0 \in S$ is an initial state, and $G \subset S$ is a finite (possibly singleton) set of goal states such that $T(s_g, a, s_g) = 1 \land C(s_g, a) = 0 \hspace{1mm} \forall a \in A, s_g \in G$.

A solution to an SSP is a policy $\pi : S \rightarrow A$ that indicates that action $\pi(s) \in A$ should be taken in state $s \in S$. A policy $\pi$ induces the value function $V^{\pi} : S \rightarrow \mathbb{R}$ that represents the expected cumulative cost $V^{\pi}(s)$ of reaching $s_g$ from state $s$ following the policy $\pi$. An optimal policy $\pi^*$ minimizes the expected cumulative cost $V^*(s_0)$ from the initial state $s_0$.

In our problem, $S = N \times \tilde{S}$ is a finite set of states comprised of the map nodes $N$ and a finite set of failure states $\tilde{S}$ and $A = E \cup \tilde{A}$ is a finite set of actions comprised of the directed edges in the graph and a finite set of recovery actions $\tilde{A}$. $T(s,a,s')$ is determined by the probability of successfully traversing the edge $e = (s,s')$, $p_e$, which is zero if the action $a$ does not correspond to the edge $e$. In a failure state $s$, $T(s, a, s') = 0$ if $a \notin \tilde{A}$ and $s \neq s'$. $C(s,a)$ is set to $t_e$ if $a \in E$, and the expected recovery cost for $s$ otherwise. Figure~\ref{fig:cpip_mdp_diagram} illustrates the planning SSP for an example urban environment.
During robot deployments, the transition function is updated to reflect the latest belief over the probability of navigation failures in traversing each edge on the map, or equivalently the probability of successful traversals. Next, we explain the method for updating the transition function.

\subsection{Updating the  Failure Belief during Deployment}

\thisWork{} builds an SSP model to represent the topological map of the environment as described in~\cref{sec:planning_background}. \thisWork{} updates the aforementioned SSP model structure during deployments as it collects more observational data from the environment, altering the underlying transition function such that the resultant model represents not just the map but the competence of the robot in traversing it.
In order to achieve that, the occurrence of a failure of class $f_i$ at edge $e$ is assumed to be a random variable from the categorical distribution $F_e \sim \text{Cat}(p_{1:L})$. The belief over this variable is defined as $\bel_t(F_e = f_{i}) = p(F_e = f_{i} | z_{1:t})$, where the subscript $t$ indicates the $t^{\text{th}}$ traversal of the edge and $z_t$ is the observation made by the robot during that traversal. Applying the Bayes rule yields
\IfConference{\vspace{-2em}}
\begin{center}
\begin{equation}
\begin{aligned}
  \belOf{t}{\f} &= \dfrac{p\left(z_t | \f, z_{1:t-1}\right) p\left(\f | z_{1:t-1}\right)}{p\left(z_t | z_{1:t-1}\right)}  \\
  &= \dfrac{p\left(\f | z_t \right) p\left(z_t \right) \belOf{t-1}{\f}} {p(\f) p(z_t | z_{1:t-1})} \quad.
\end{aligned}
\end{equation}
\end{center}
Defining the negation of $\f$ as $p(\neg\f) = 1 - p(\f) = \sum_{j\neq i}p(f_{j,e})$ the belief can be implemented as the log odds ratio
\IfConference{\vspace{-2em}}\IfJournal{\vspace{-2em}}
\begin{center}
\begin{equation}
\begin{aligned}\label{eq:log_odds}
  l_t(\f) &= \log\left( \dfrac{\belOf{t}{\f}} {\belOf{t}{\neg\f} } \right) \\
          &= \log\left( \dfrac{p(\f | z_t)}{1 - p(\f | z_t)} \right) 
         + l_{t-1}(\f) - l_0(\f)  \quad,
\end{aligned}
\end{equation}
\end{center}
where $l_0(\f) = \log\left( \frac{p(\f)}{1 - p(\f)} \right)$ is the prior in log odds form.
Before the first deployment of the robot in a new environment, $p(\f) = \epsilon$ and $p_e = p(f_{L,e}) = 1 - \sum_{i \neq L}p(\f)$ for every $e \in E$.
Then upon each traversal of an edge, the above relation is used to update the transition function of the planning SSP model such that $T_t(s, e, \tilde{s_i}) = \belOf{t}{\f} = 1 - \frac{1}{1 + \exp(l_t(\f))}$.
The main term that needs to be computed for updating the belief in Eq.~\ref{eq:log_odds} after each traversal is $p\left(\f | z_t \right)$, which is known as the inverse observation likelihood and in \thisWork{} it is implemented by two different functions, each handling one of the two different types of observations $z_t$:
1) Occurrence of failures of class $f_i$ which is indicated via intervention signals issued either by a human or a supervisory sensing unit and is denoted by $s_{t,i}$; 2) Sensory input that the robot continuously acquires such as RGB images captured by cameras on the robot, which is denoted by $\mathbf{I}_t$.
For the former, the inverse observation likelihood is implemented as
\vspace{-3em}
\begin{center}
\begin{align}
    p\left(\f | z_t = s_{t,j} \right) =
     \begin{cases}
                 \delta                   & i = j \\
                 \frac{1 -\delta}{L - 1}  & i \neq j
     \end{cases} 
\end{align}
\end{center}
where $\delta$ is a constant coefficient. 
The inverse observation likelihood function for the latter type of observations, however, is machine learned and is one of the key components of this work that allows \thisWork{} to reach an accurate estimate of $\belOf{t}{\f}$ without requiring the robot to experience costly failures. 
\thisWork{} structures the problem of learning $p(\f | z_t = \mathbf{I}_t)$ such that it can be achieved with a small number of failure examples for training data. Introspective perception is leveraged to extract features associated with errors in perception from the high dimensional raw sensory data. These features are then used to learn to predict the probability of different classes of failures of navigation. By learning this likelihood function, the robot will learn to better navigate its environment, proactively avoiding paths that are known to lead to failure cases, and reactively adjusting its policy upon encountering novel situations that may lead to failures. In the following section we describe the different parts of this learning problem.

\section{Failure Prediction via Introspective Perception}\label{sec:introspection}

\begin{figure*}[t]
  \centering 
  \vspace{2mm}
  \includegraphics[width=0.9\linewidth ,trim=0 0 0 0,clip]{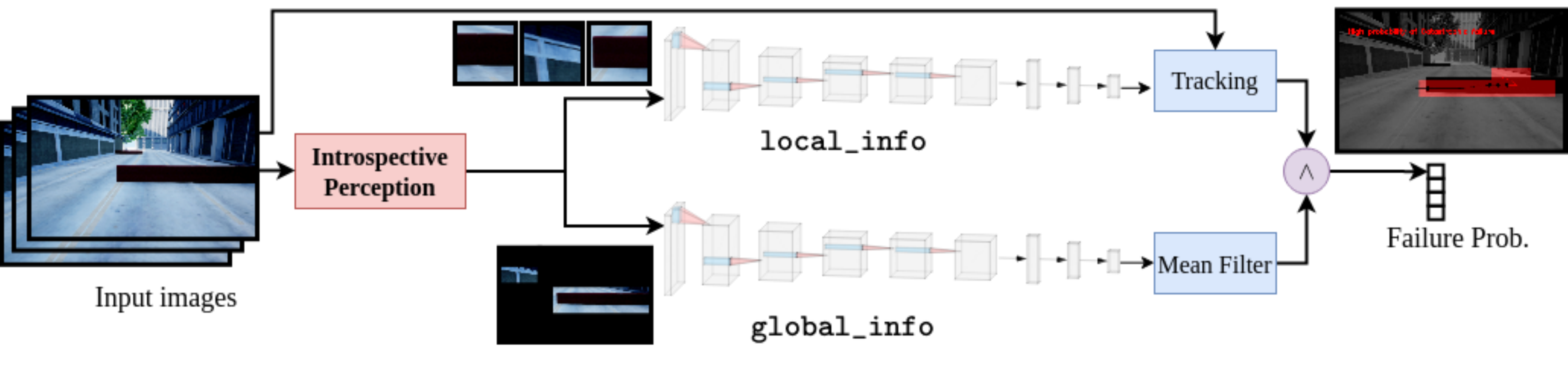}
  \caption{Navigation competence predictor model architecture.}
  \label{fig:comp_pred_model}
  \vspace{-1mm}
\end{figure*}

In order to predict failures of navigation given the sensory data, we need to approximate the function $p(\f | z_t = \mathbf{I}_t)$.
End-to-end learning of this function is intractable because it requires a great amount of training data, yet catastrophic failures in robotics when executing tasks such as autonomous navigation do not happen frequently. The scarcity of these examples makes it challenging to learn a classifier that predicts the probability of task execution failure directly from the raw sensory data. Without enough training data and without abstracting the acquired high dimensional sensory data, the learned classifier is bound to overfit to the training data. We instead propose to factorize $p(\f | z_t) = \int_{\phi} p(\f | \phi)p(\phi | z_t)$ where $\phi$ are the features extracted from observations by introspective perception --- a model-free approach to predicting arbitrary errors of perception.

\subsection{Introspective Perception}
Early works on introspective perception~\cite{zhang2014predicting, daftry2016} defined a perception algorithm to be introspective if it is capable of predicting the occurrence of errors in its output given the current state of the robot. Follow-up works~\cite{rabiee2019ivoa, rabiee2020ivslam} extended this definition and required such perception algorithms to predict the probability of perception error conditioned on the region in the raw sensory data that the output is dependent upon, e.g. an image patch in the captured image by an RGB camera where the estimated depth of the scene is erroneous. This is obtained by means of an introspection function that is trained on empirical data. 

In \thisWork{}, the robot is assumed to be equipped with one or more introspective perception modules; each module has a learned function $\mathbb{I}_k: Z \rightarrow \mathbb{R}^n$, which extracts features $\phi \in \mathbb{R}^n$ from the raw sensory data $Z$ that encode information about sources of perception errors. The outputs of all introspective perception modules are fed to a navigation competence predictor $\compModel{}: \mathbb{R}^{n \times K} \rightarrow [0, 1]^{L}$, which learns to estimate the likelihood of each of the different classes of failure ${f}_{1:L}$ given a set of sources of perception errors, i.e. $P(f_l | \phi_{1:K})$ such that $\sum_l P(f_l | \phi_{1:K}) \leq 1$. The inverse observation likelihood function in Eq.~\ref{eq:log_odds} is then estimated as the composition of the above two functions, i.e. 
$p(\f | z_t = \mathbf{I}_t) = \compModel{}\left(\mathbb{I}_{1:k} \left( \mathbf{I}_t \right) \right)_{[i]}$.
\changes{It should be noted that although \thisWork{} assumes a constant set of failure classes, the number of distinct sources of failures, which is often much larger than the number of different classes of failure, are not enumerated \emph{a priori}. Each failure class corresponds to a different severity level and hence a different failure recovery cost that is considered in planning. There exist, however, a large number of failure sources that lead to failures with the same severity level. 
For example, a high-severity class of failure in robot navigation are collisions, for which there exist numerous failure sources including false negatives in obstacle detection due to texture-less surfaces or small object sizes, terrain type mis-classification, dynamic obstacles, occlusions, etc.
Furthermore, while the distinct sources of failure differ between environments, the classes of failure are specific to the objective of the domain (e.g. navigation, manipulation, etc.), irrespective of the environment, and are hence comparatively easy to enumerate.}

In this paper we implement introspective perception for a block matching-based stereo depth estimator~\cite{pulli2012real} using the same convolutional neural network architecture as that used in~\cite{rabiee2019ivoa} for the introspection function. The training data is collected autonomously using a depth camera as supervisory sensing, which is only occasionally available and provides oracular information about the true depth of the scene.

\subsection{Competence Predictor Model}\label{sec:competence_pred_model}

We implement the navigation competence predictor model $\compModel{}:\mathbb{R}^{n \times K} \rightarrow [0, 1]^{L}$ as an ensemble of two deep neural networks. The input to the model is a list of image patches $I_i \in {R}^{n}$ extracted from the same input image $\mathbf{I}$ and the output is the probability of each class of failure. The architecture as shown in Figure~\ref{fig:comp_pred_model} consists of two sub-models that are trained independently.

The \texttt{global\_info} network is a convolutional neural network (CNN) that operates simultaneously on all input image patches arranged on a blank image in their original pixel coordinates. The input to this network is equivalent to the input image masked at all regions except for those predicted by introspective perception to lead to errors. The \texttt{global\_info} CNN captures task-contextual and spatial information from the current frame related to competence. By masking out parts of the full image deemed to be unrelated to perception failures we are able to ensure that the \texttt{global\_info} CNN does not overfit to specific environments.

The \texttt{local\_info} network is a CNN that is fed as input individual image patches. The output of this branch is the probability of each class of failure for each single image patch. 
This network learns correlations between navigation failures and image features that lead to perception errors.
The goal of this branch is to locally pinpoint the potential source of navigation failures in the image space, when a class of failure is predicted by the \texttt{global\_info} network.

The last stage of the model is a temporal filtering of the output of each of the two networks. Failure class probabilities that are produced by the \texttt{global\_info} network are passed through a mean filter to output $P_f \in [0, 1]^{L}$. Moreover, image patches that are predicted by the \texttt{local\_info} network to lead to navigation failures are tracked in the full image over consecutive frames to form a set of active tracklets $\Lambda_i$ for each class of failure $f_i$.
The output of the model is obtained via strict consensus on the output
such that 
\begin{equation}
 P(f_i) =
 \begin{cases}
             {P_f}_{[i]} & \text{if $\Lambda_i \neq \emptyset$} \\
             0 & \text{otherwise}
 \end{cases} 
\label{eq:comp_model}
\end{equation}
In other words, the predicted probability of each class of failure provided by the \texttt{global\_info} network is only accepted if the \texttt{local\_info} network also supports that by detecting at least one potential cause for the same class of failure in the image space. During deployment, if $ \exists j \mid P(f_j) > \epsilon $, i.e. there exists consensus between the two branches of the network on the existence of any class of failure, the output of the competence predictor model will be used to update the belief in Eq.~\ref{eq:log_odds}.

\vspace{-1em}
\changes{
\section{Implementation Details} \label{sec:impl_details}
In this section we provide implementation details for our application of \thisWork{}, i.e. path planning for an unmanned ground vehicle (UGV) that uses a stereo vision-based depth estimator~\cite{pulli2012real} for obstacle avoidance.
}
\changes{
\subsection{Autonomy Stack}
Our navigation software consists of global path planning on the navigation graph and local path planning to follow the planned path while avoiding dynamic obstacles that the robot does not know about \emph{a priori}. \thisWork{}'s planning model does the global path planning and we use a trajectory roll-out local path planner. The 3D reconstruction of the environment by stereo-vision is processed and any points with their height larger than \SI{15}{\cm} and less than the height of the robot are detected as obstacles. All obstacle points coordinates are projected to the ground plane, converted to a 2D laser-scan format, and used by the local path planner to select the least cost trajectory from a set of sampled trajectories, such that the robot keeps a large clearance from obstacles and makes progress towards the next way-point on the global plan.}

\changes{
We implement introspective perception for the depth estimator and train a CNN to predict depth estimation errors similar to prior work~\cite{rabiee2019ivoa}. The network is composed of 5 convolution layers followed by 3 fully connected layers. The input to the network are the image patches of size $70 \times 70$ pixels extracted from the $512 \times 512$ pixel images captured by the left stereo pair. The output is the probability of depth estimation error and all image patches predicted to lead to perception error with a probability of $>0.5$ are passed to the competence predictor model. We use a similar CNN architecture as that use for introspective depth estimation for both sub-models of the competence predictor with the only difference being the number of nodes in the output layer of the network.
The full pipeline of navigation competence predictor which consists of introspective perception and the competence predictor model runs at \SI{5}{\hertz} on a laptop with Intel Core i9-9880H and GeForce RTX 2080 Max-Q.}

\subsection{Training of \thisWork{}}
\thisWork{} has two learned components, i.e. introspective perception and the competence predictor model and they are trained sequentially. The training data is extracted from logs of robot deployments in the training environment. The logs include data collected by the primary sensors, i.e. RGB images captured by the stereo cameras, as well as data collected by supervisory sensors, i.e. the Orbbec Astra depth camera that is only occasionally available. Furthermore, intervention signals issued by a human operator upon occurrence of navigation failures are also recorded. 

The deployment logs are processed offline. First, introspective perception is trained with data that is autonomously labeled using the supervisory sensing.
Then, the training data for the competence predictor model is prepared by
passing the raw sensory data through the introspective perception module and labeling the output image patches as associated with one of the classes of navigation failures if they fall within a fixed time window preceding the occurrence of such failures.
Each of the two sub-models of the competence predictor model explained in \S\ref{sec:competence_pred_model} are then trained using a cross-entropy loss.

%% file: experimental_results.tex
\section{Experimental Results} \label{experimental_results}

In this section: 
1) We evaluate \thisWork{} on how well it predicts sources of robot failures.
2) We compare \thisWork{} against baseline global path planners in terms of their task completion success rate and their task completion time.
3) We evaluate the importance of introspective perception in \thisWork{}'s performance and generalizability via ablation studies.

\subsection{Experimental Setup}\label{sec:experiment_setup}

\begin{figure}[t]
\centering
\vspace{2mm}
\includegraphics[width=1.0\linewidth]{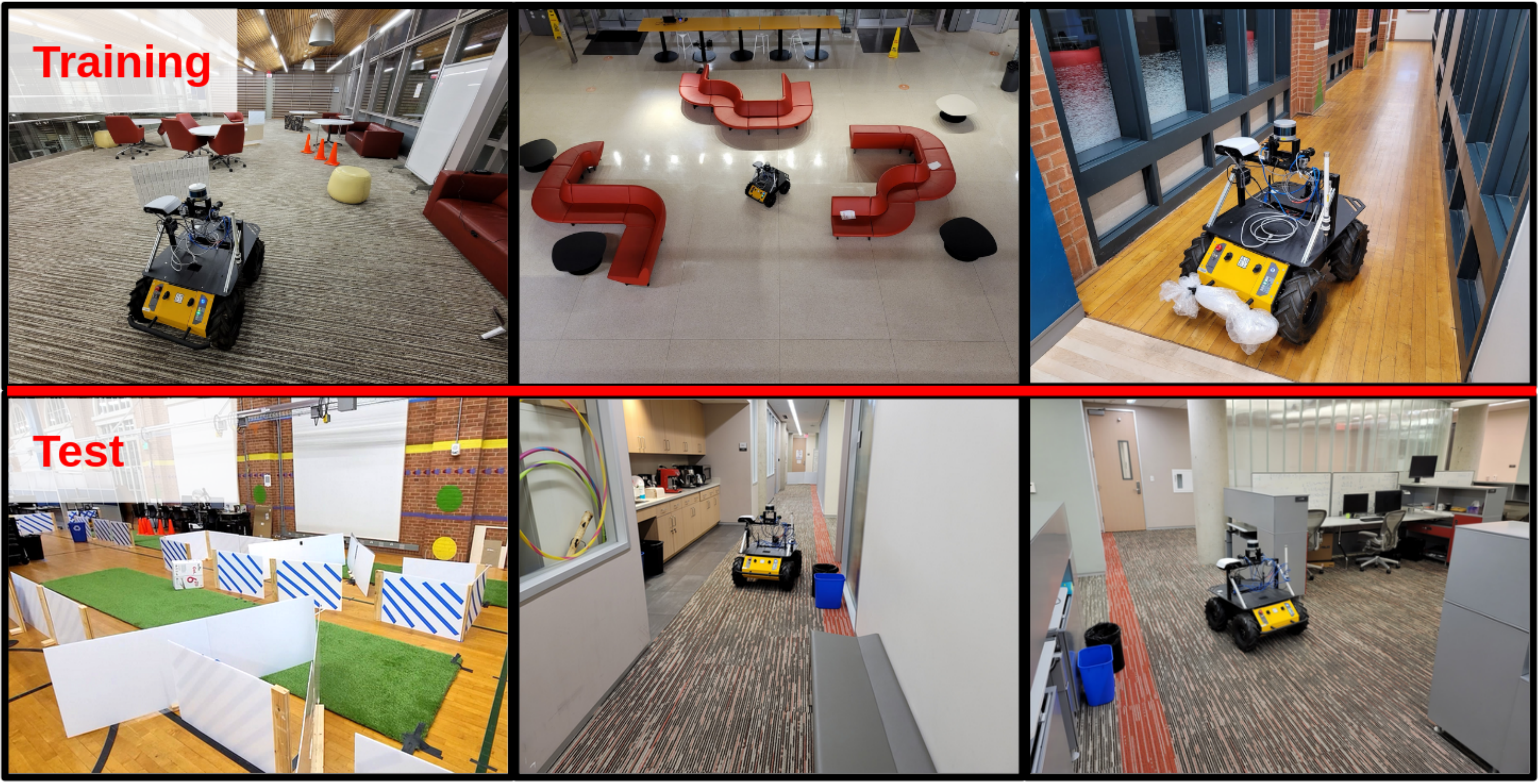}
\caption{Training and test environments in the real-world experiments.}
\label{fig:robot_exp_env}
\vspace{-1.5em}
\end{figure}

\paragraph{Simulation}
In order to evaluate \thisWork{} and compare it against SOTA extensively, we use AirSim~\cite{shah2018airsim}, a photo-realistic simulation environment, where robot failures are not expensive and the robot can easily be reset upon occurrence of navigation failures. A simulated car is equipped with a stereo pair of RGB cameras as well as a depth camera that provides ground truth depth readings for every pixel in the camera frame. We use two separate urban environments for training and testing. The environments are populated with obstacles of different shapes, textures, and colors.
\changes{
\paragraph{Real-robot maze}
We also evaluate \thisWork{} on a real robot. We use a Clearpath Husky robot equipped with a stereo pair of RGB cameras, an Orbbec Astra depth camera, and a Velodyne VLP-16 3D Lidar. We use different indoor sites for training and testing of \thisWork{}. Each environment has different types of terrain such as tiles and carpet, and is populated with obstacles of different shapes, textures, and surface materials. The test environment is a maze constructed in an area of size \SI{60}{\meter^2}.
}
\changes{
\paragraph{Real-robot large-scale}
In order to test \thisWork{} extensively and in more natural settings, we also conduct a large scale experiment, where we deploy the robot on the entire floor of a building. This test environment has an area of larger than \SI{400}{\meter^2} and the robot traverses more than \SI{1.5}{\kilo\meter} during the deployment. 
Figure~\ref{fig:robot_exp_env} shows the training and both the large scale and maze test environments.
}

\subsection{Failure Prediction Accuracy}
\label{sec:failure_pred_acc}

\begin{figure}[t]
\centering
\includegraphics[width=0.6\linewidth]{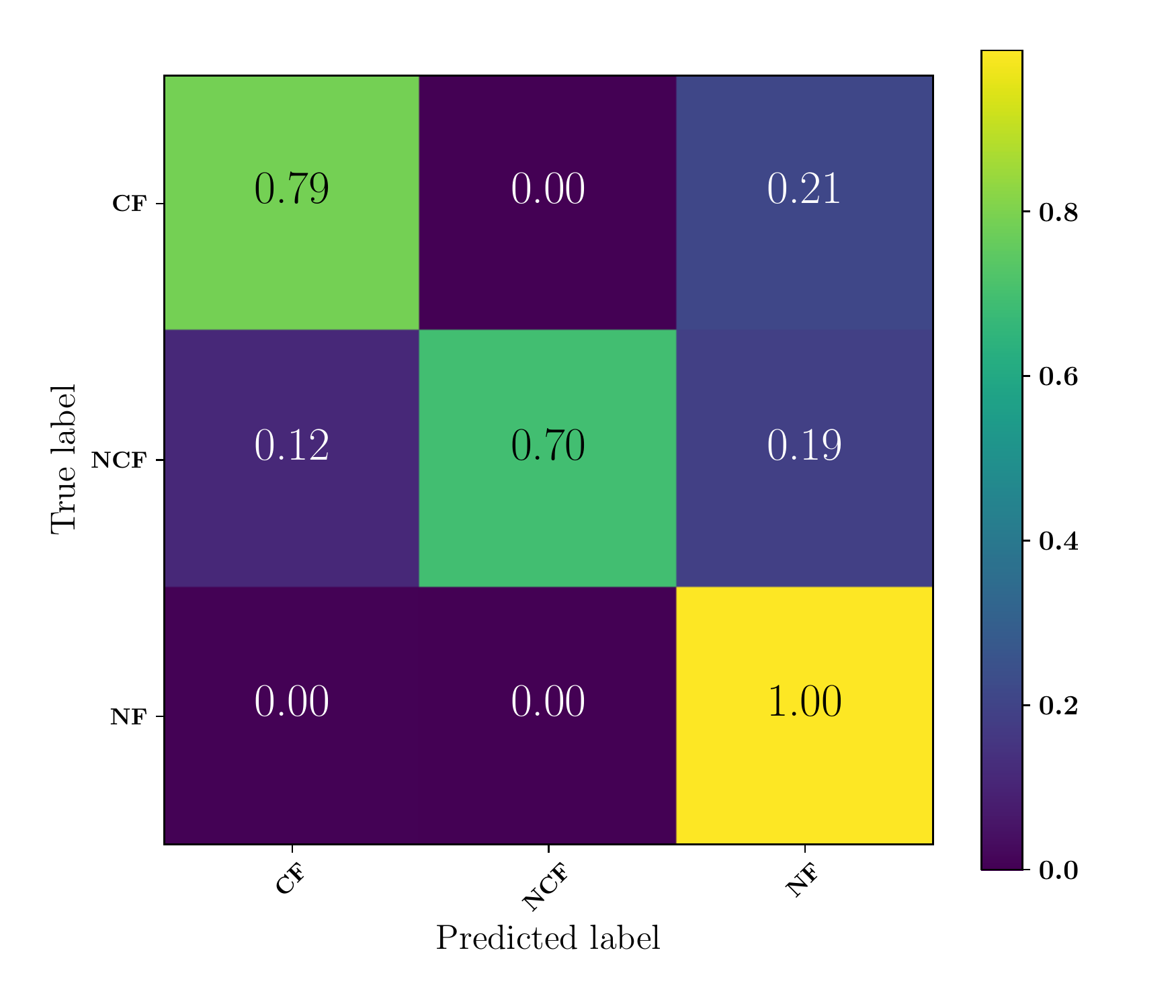}
\caption{Prediction results of the competence predictor model in the previously unseen test environment for the three classes of catastrophic failures (CF), non-catasatrophic failures (NCF), and no failures (NF).
}\label{fig:comp_predictor_conf_matrix}
\IfConference{\vspace{-1em}}
\end{figure}

In order to evaluate the accuracy of \thisWork{} in predicting failures of navigation, we have the autonomous agent traverse each of the edges of the navigation graph in the test simulation environment 50 times and run the captured images by the robot camera through the \thisWork{}'s introspective perception module and the competence predictor model to predict instances of navigation failure.
In this paper, we implement \thisWork{} with two classes of failures.
\begin{inparaenum}[1)]
\item Catastrophic failures, where the robot ends up in a state that precludes completion of the task and is not recoverable with human intervention. Examples of this class include collisions and the robot getting stuck off-road.
\item Non-catastrophic failures, where the robot will not be able to complete its task unless intervention is provided by a human operator or a supervisory sensor. The robot getting stuck due to false detection of obstacles or because of localization errors are examples of this type of failure. 
\end{inparaenum}
Figure~\ref{fig:comp_predictor_conf_matrix} illustrates the predicted and actual navigation failures in a confusion matrix. \thisWork{} correctly predicts occurrence of navigation failures more than $70\%$ of the time for both types of failures. Prediction errors mostly correspond to cases, where the source of failure is significantly different looking from the examples available in the training data.

\begin{figure}[t]
\centering
 \begin{subfigure}[b]{0.49\linewidth}
  \includegraphics[width=1.0\linewidth, trim=0 0 0 0,clip]{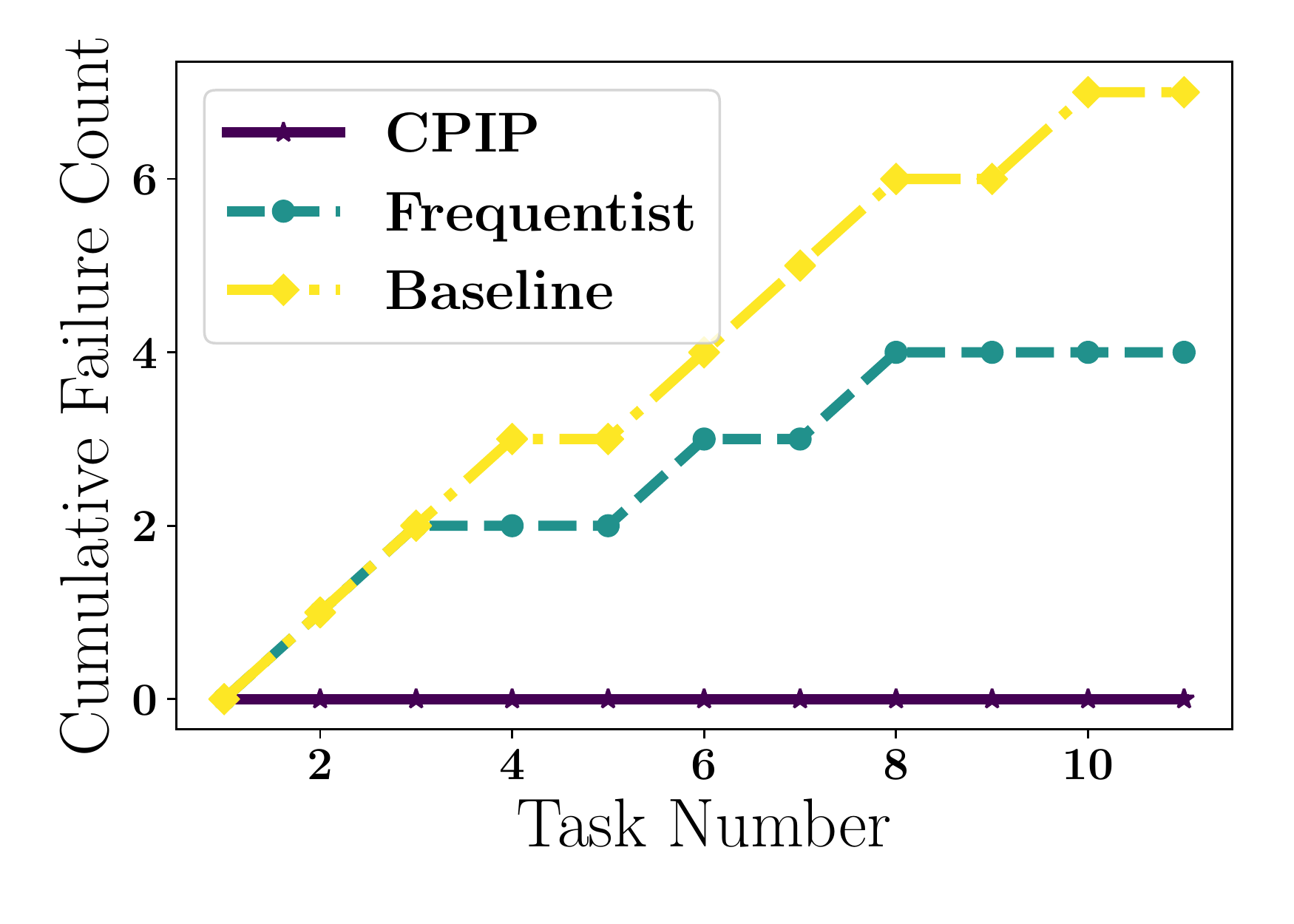}
 	\caption{}
 	\label{fig:failure_cum_robot}
 \end{subfigure}
 \begin{subfigure}[b]{0.49\linewidth}
  \includegraphics[width=1.0\linewidth, trim=0 0 0 0,clip]{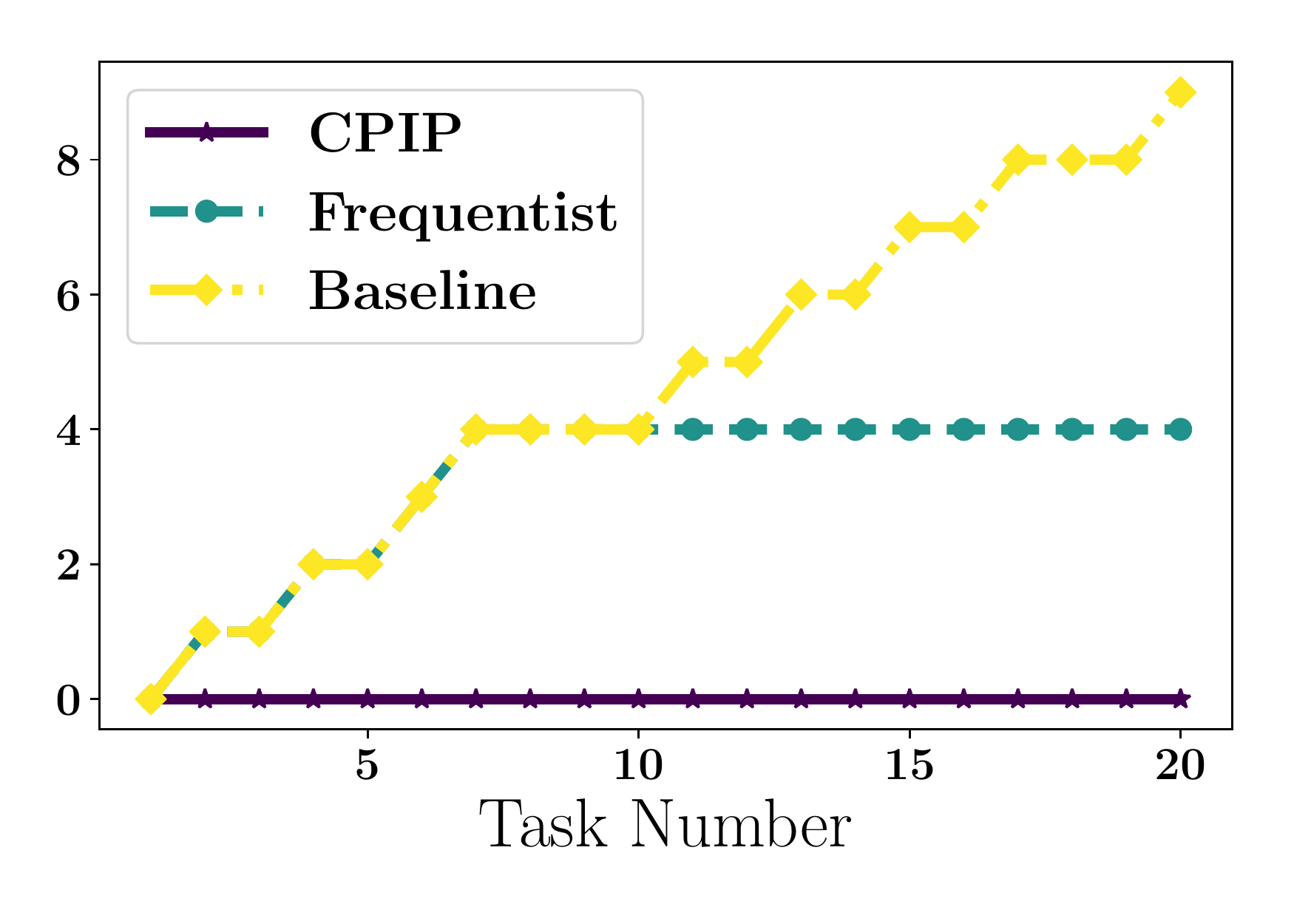}
 	\caption{}
 	\label{fig:failure_cum_robot_ext}
 \end{subfigure}
  \caption{\textcolor{\changesColor}{Comparison of cumulative failure count (\subref{fig:failure_cum_robot}) in the real-robot maze experiment, and (\subref{fig:failure_cum_robot_ext}) the real-robot large scale experiment for this work (\thisWork{}), SOTA (frequentist), and the baseline with no competence-aware planning.}}
 \label{fig:failure_count_vs_time}
\end{figure}

\begin{figure*}[h]
\centering
 \includegraphics[width=0.96\linewidth, trim=10 0 10 0,clip]{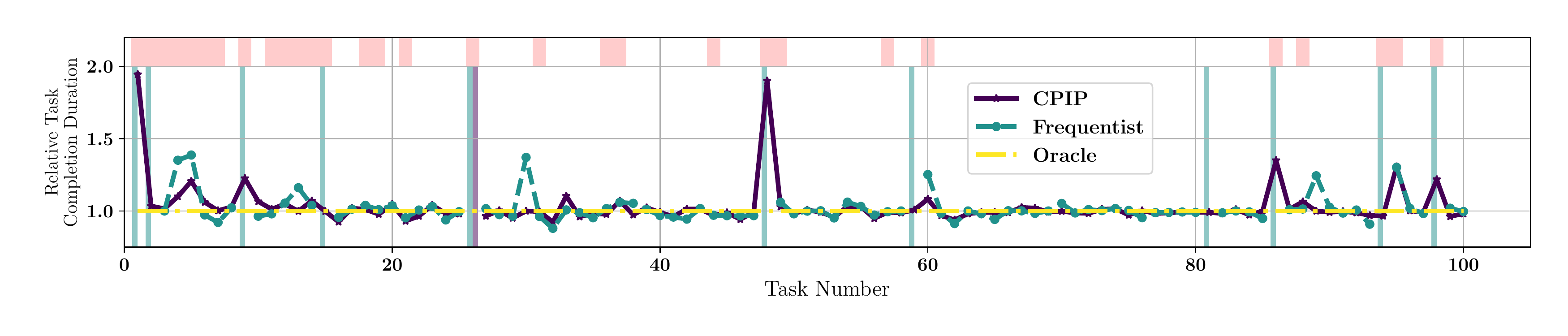}
 \caption{\textcolor{\changesColor}{Task completion duration w.r.t. an oracle planner that is provided with the true probability of failure throughout the environment ahead of deployment. 
  Vertical bars visualize the incomplete tasks for each method annotated by color.
  Highlighted red regions in the top band demonstrate tasks, during which the robot encounters previously unseen parts of the environment. Best viewed in color.}}
  \label{fig:task_completion_duration}
\end{figure*}

\subsection{Navigation Success Rate and Plan Optimality}

\begin{table}[t]
\caption{\textcolor{\changesColor}{\textsc{Task completion and failure prevention rate.}}}
\label{tab:result_summary}
\centering
\resizebox{\linewidth}{!}{%
\begin{threeparttable}
\begin{tabular}{@{}llccllcc@{}}
\toprule
                                                                                                   &             & \multicolumn{1}{l}{\multirow{2}{*}{\# Tasks}} & \multirow{2}{*}{TCR (\%)} & \multicolumn{2}{l}{Relative TCD\tnote{*}} & \multicolumn{2}{c}{\# Avoided Failures} \\ \cmidrule(r{0.2em}){5-6} \cmidrule(l{0.2em}){7-8}
                                                                                                   &             & \multicolumn{1}{l}{}                          &                           & Mean                & Std        & CF                 & NCF                \\ \midrule
\multirow{2}{*}{\begin{tabular}[c]{@{}l@{}}Real Robot\\ Maze\end{tabular}}                                                    & CPIP        & 11                                            & \textbf{100}              & \textbf{1.04}       & 0.08       & \textbf{5 (100\%)} & \textbf{3 (100\%)} \\
                                                                               & Frequentist & 11                                            & 73                        & 1.22                & 0.43       & 3 (60\%)           & 1 (33\%)           \\ \midrule
\multirow{2}{*}{\begin{tabular}[c]{@{}l@{}}Real Robot\\ Large-Scale\end{tabular}} & CPIP        & 20                                            & \textbf{100}              & 0.98                & 0.04       & \textbf{9 (100\%)} & \textbf{-}         \\
                                                                               & Frequentist & 20                                            & 80                        & 0.99                & 0.04       & 5 (55\%)           & \textbf{-}         \\ \midrule
\multirow{2}{*}{Simulation}                                                    & CPIP        & 100                                           & \textbf{97}               & 1.00                & 0.05       & \textbf{14 (93\%)} & \textbf{61 (97\%)} \\
                                                                               & Frequentist & 100                                           & 83                        & 1.02                & 0.09       & 9 (60\%)           & 52 (83\%)          \\ \bottomrule
\end{tabular}
\begin{tablenotes}
\item[*] Task completion duration statistics are only calculated for tasks that were completed by both algorithms.
\end{tablenotes}
\end{threeparttable}
}
\vspace{-1em}
\end{table}

\begin{figure*}[t]
\centering
 \begin{subfigure}[b]{0.365\linewidth}
  \includegraphics[width=0.95\linewidth, trim=0 0 0 0,clip, right]{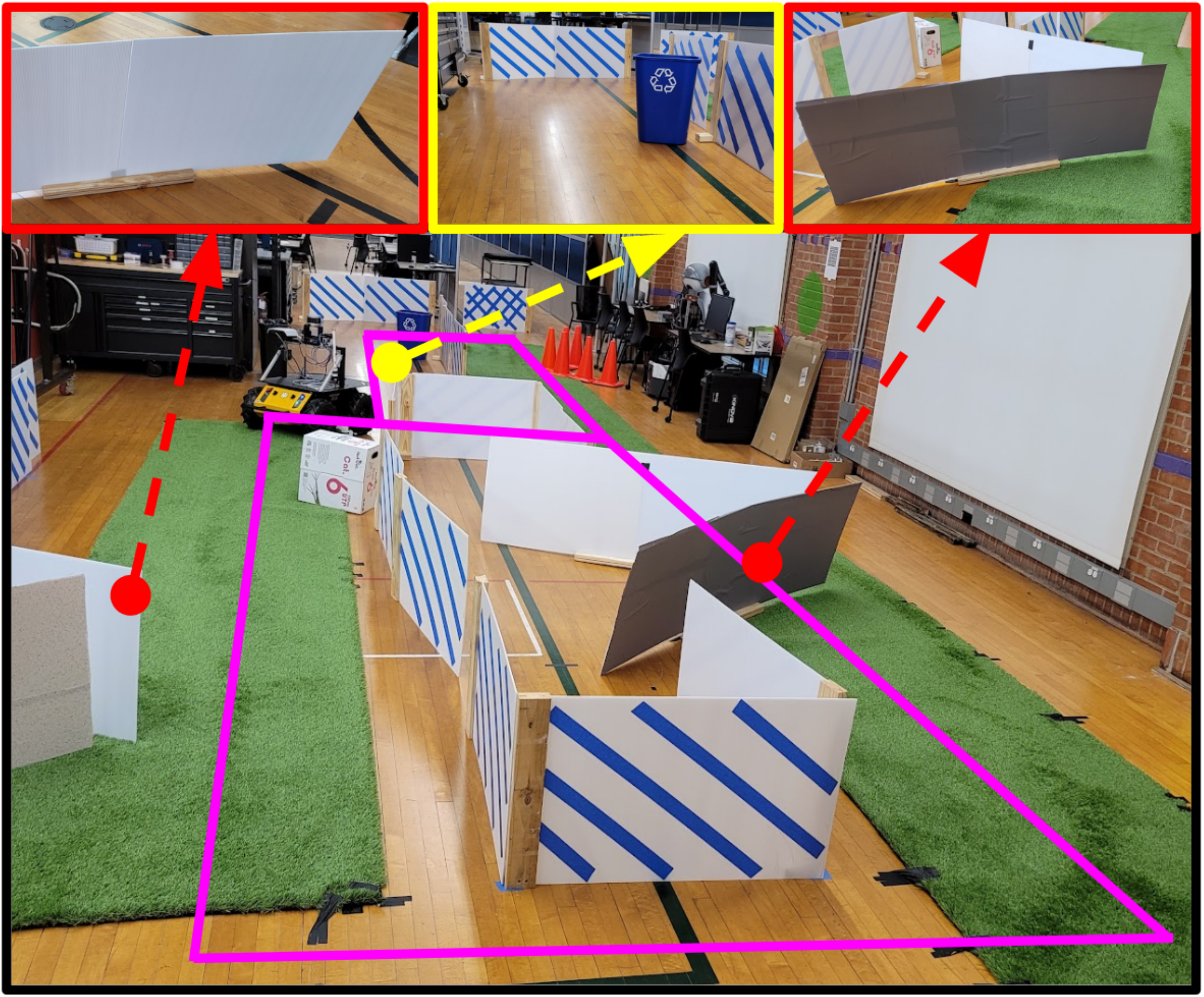}
 	\caption{}
 	\label{fig:robot_exp_snapshots}
 \end{subfigure} \hspace*{-1.2em}
 \begin{subfigure}[b]{0.33\linewidth}
  \includegraphics[width=0.95\linewidth, trim=0 0 0 0,clip, right]{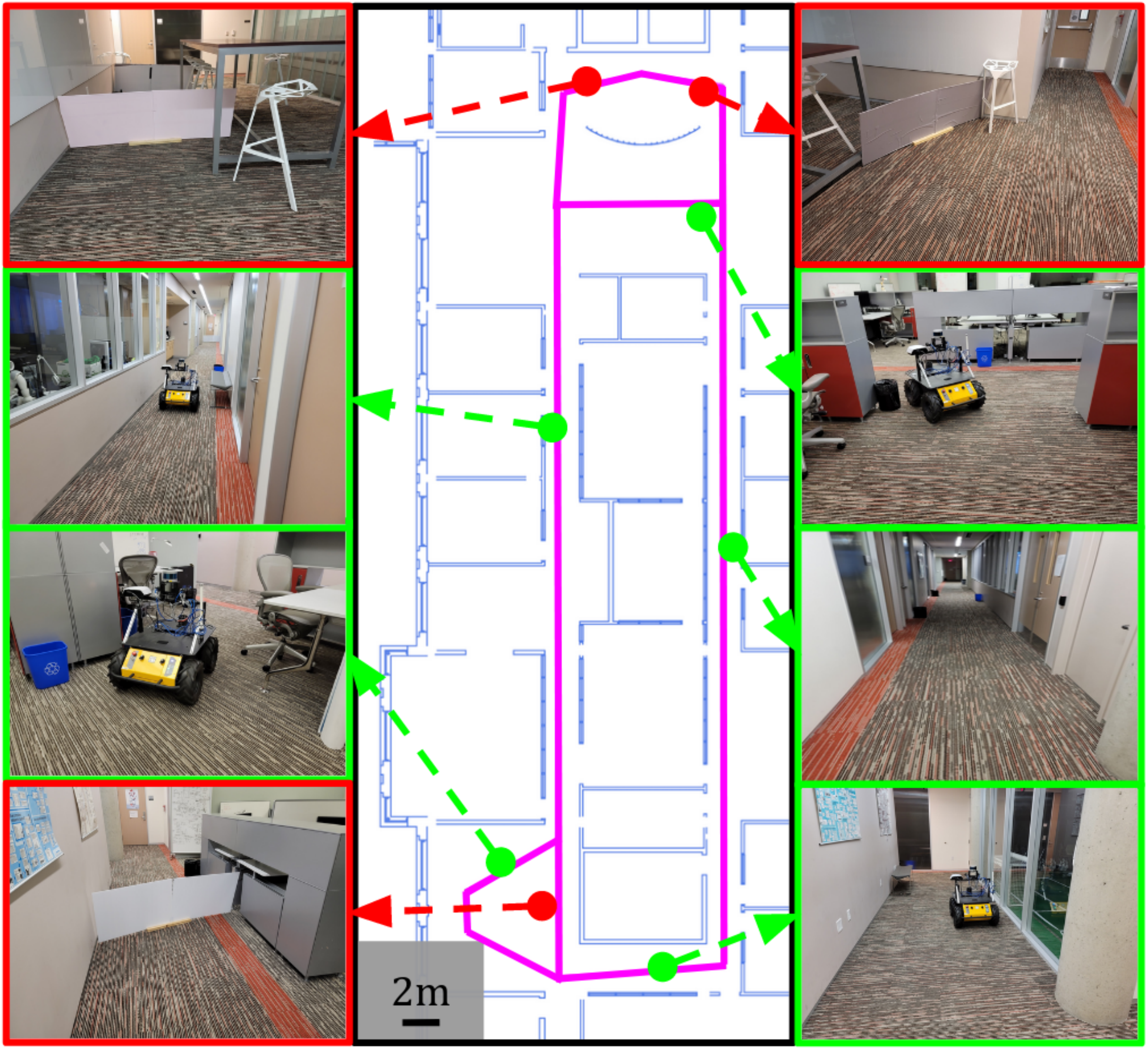}
 	\caption{}
 	\label{fig:robot_exp_ext_snapshots}
 \end{subfigure}
 \begin{subfigure}[b]{0.29\linewidth}
  \includegraphics[width=0.95\linewidth, trim=0 0 0 0,clip, left]{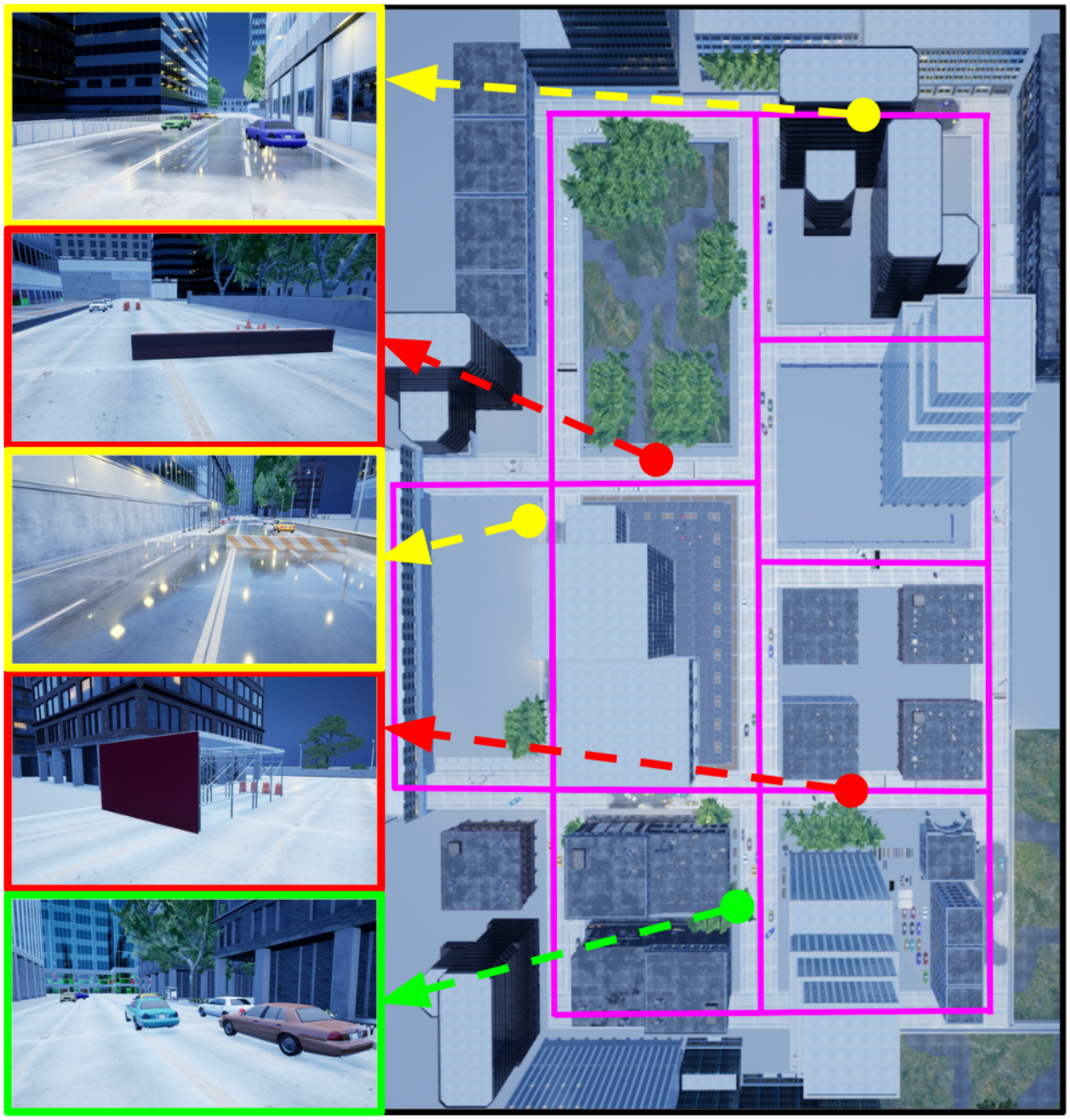}
 	\caption{}
 	\label{fig:sim_exp_snapshots}
 \end{subfigure} 
 \caption{\textcolor{\changesColor}{Test environments in~(\subref{fig:robot_exp_snapshots}) the real robot maze experiment,~(\subref{fig:robot_exp_ext_snapshots}) large-scale real robot experiment, and~(\subref{fig:sim_exp_snapshots}) the simulation experiment. Regions of the environments highlighted in red cause catastrophic failures, regions highlighted in yellow illustrate sources of non-catastrophic failures, and areas annotated with green, show areas where the robot can successfully operate autonomously.}}
 \label{fig:test_environments_2}
 \IfConference{\vspace{-1.5em}}
\end{figure*}

We test the end-to-end system in predicting navigation failures and leveraging this information to proactively plan paths that reduce the probability of failures, by deploying the robot in a previously unseen test environments. 
The robot is commanded to complete randomly generated navigation tasks that consist of a starting pose and a target pose.
We conduct this experiment in all three settings explained in~\cref{sec:experiment_setup}, i.e. simulation, real-robot maze, and the large-scale real-robot deployment.

We compare \thisWork{} with a baseline path planner that does not reason about the competence of the robot as well as a state-of-the-art approach for competence-aware path planning --- called the Frequentist approach --- that relies on keeping track of the frequency of past failures in traversing each of the edges of the navigation graph~\cite{lacerda2014optimal}. 
Figure~\ref{fig:failure_count_vs_time} compares the cumulative failure count for all three methods throughout the real-robot experiments. With the Frequentist approach, the robot learns to avoid regions of the environment, where it cannot navigate reliably as it experiences navigation failures. However, \thisWork{} enables the robot to predict and avoid most of these failures, leading to the least number of experienced failures. 

We also evaluate the optimality of the planned paths by comparing the task completion time for all the methods under test with an oracular path planner that is given the true probability of navigation failures for each edge of the navigation graph. The ground truth failure probabilities are obtained by having the agent traverse each edge of the navigation graph numerous times and logging the frequency of each type of failure.
Figure~\ref{fig:task_completion_duration} shows the completion duration for each task in the simulation experiment. The duration values are normalized by the task completion duration when the oracular path planner is used. The figure also illustrates instances of task completion failures for both \thisWork{} and the Frequentist method. Such instances include occurrence of catastrophic failures or occurrence of consecutive non-catastrophic failures such that the robot cannot recover from a stuck state by re-planning. \thisWork{} task completion duration is similar to that of the oracular path planner except for tasks where the robot visits a previously unseen part of the environment and has to re-plan upon prediction of a source of navigation failure. An example of such re-planning can be seen around task number 50 in Figure~\ref{fig:task_completion_duration}.
\changes{Table~\ref{tab:result_summary} summarizes the task completion rate (TCR), task completion duration (TCD), and the number of avoided navigation failures by \thisWork{} and the Frequentist method for both simulation and real-robot experiments.
\thisWork{} achieves a significantly higher TCR across all experiments; moreover, CPIP either performs similarly or outperforms the Frequentist approach in terms of TCD. The reduced task completion duration achieved by CPIP is due to proactively predicting and avoiding non-catastrophic failures, e.g. getting stuck behind falsely detected obstacles. The Frequentist approach, on the other hand, would experience these failures and although it might be able to eventually complete the task by replanning, it will suffer from a longer task completion duration. This effect was more pronounced in the real-robot maze experiment, where the distance traveled by the robot in each task was shorter compared to the other experiments, hence the relative task completion delay caused by non-catastrophic failures was larger. }
Figure~\ref{fig:test_environments_2} illustrates snapshots of the test environments and highlights the different sources of navigation failures encountered by the robot, which includes different types of texture-less obstacles as well as reflective surfaces.

\subsection{Ablation Study}

\begin{figure}[h]
\centering
 \begin{subfigure}[b]{0.45\linewidth}
  \includegraphics[width=1.0\linewidth, trim=0 0 0 0,clip]{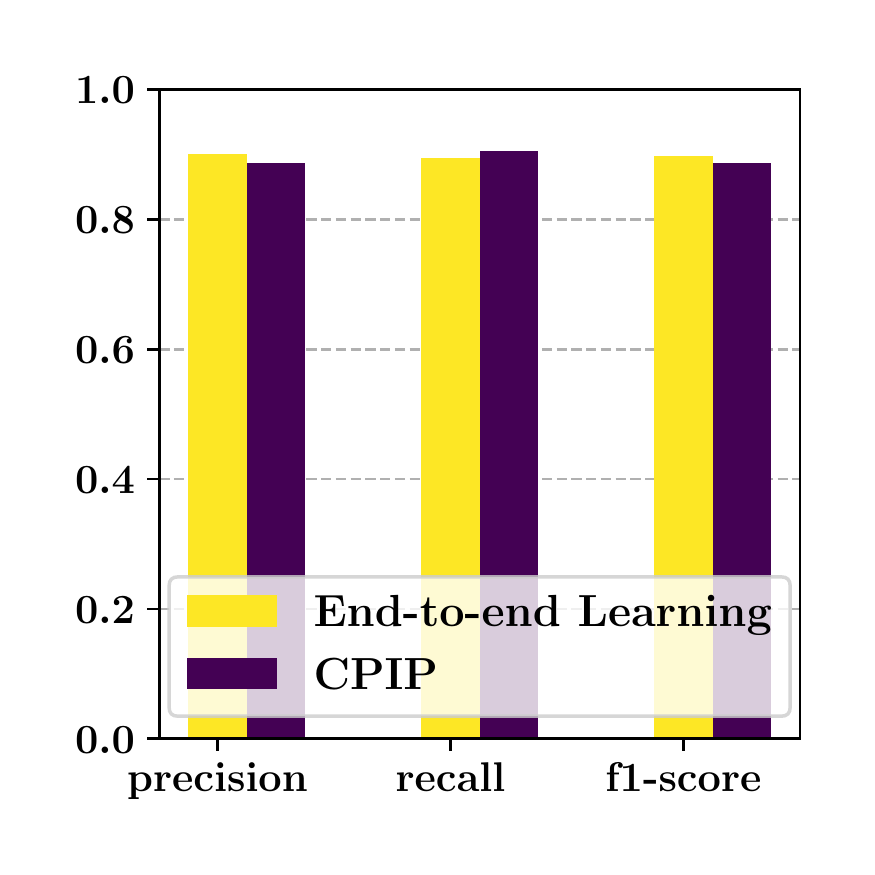}
 	\caption{Known environment}
 	\label{fig:seen_test_env}
 \end{subfigure}
 \begin{subfigure}[b]{0.45\linewidth}
  \includegraphics[width=1.0\linewidth, trim=0 0 0 0,clip]{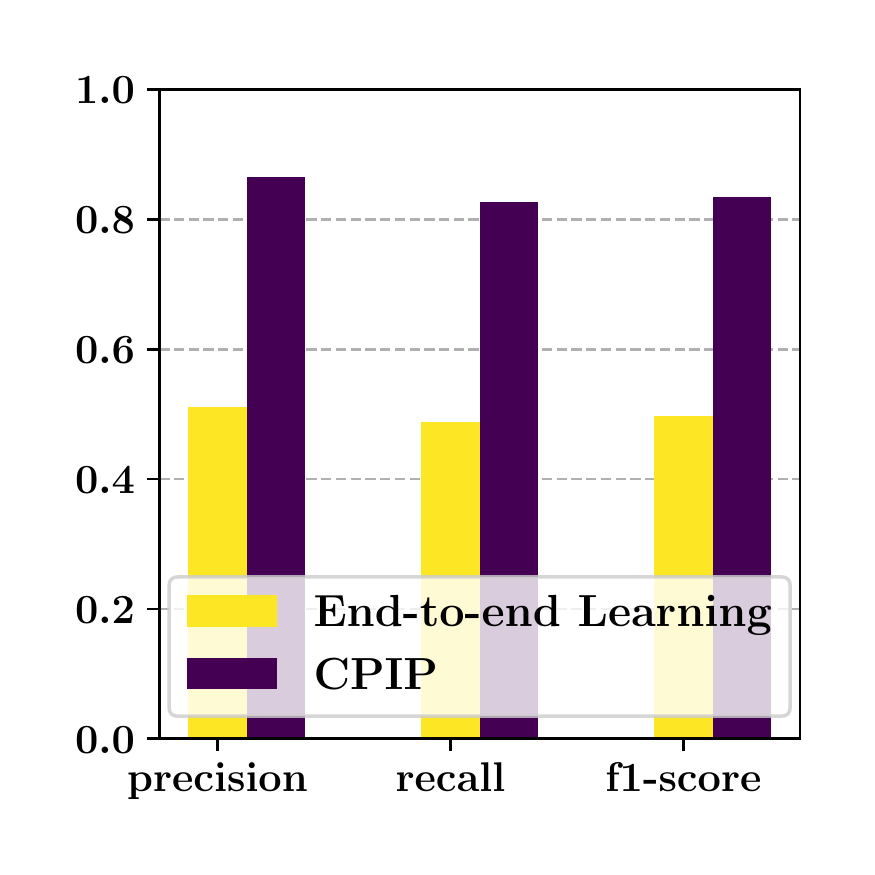}
 	\caption{Novel environment}
 	\label{fig:unseen_test_env}
 \end{subfigure}
  \caption{Results of the navigation failure prediction for \thisWork{} vs. an end-to-end classifier that does not use introspective perception (\subref{fig:seen_test_env}) in a previously seen environment and (\subref{fig:unseen_test_env}) in a novel environment.}
 \label{fig:classification_report}
 \vspace{-1.5em}
\end{figure}
In order to evaluate the importance of introspective perception in the pipeline of \thisWork{}, we conduct an ablation study.
We train a classifier that instead of leveraging the extracted information by introspective perception, directly receives the raw captured RGB images as input and outputs the probability of each class of failure occurring in a specified time window in the future. We use a convolutional neural network with the AlexNet architecture similar to that used in prior work~\cite{daftry2016} for predicting failures of perception.

We train the classifier on the same simulation dataset used for training \thisWork{} and we compare the performance of both methods in predicting navigation failures both in a previously unseen environment---the same test dataset described in ~\cref{sec:failure_pred_acc}---as well as in a new set of deployments of the agent in the training environment. 
Figure~\ref{fig:classification_report} shows the average precision, recall, and f1-score metrics over all classes, i.e. two classes of failures and a no-failure class, for both \thisWork{} and the end-to-end classifier.
While both methods perform similarly good in a previously seen environment, 
\thisWork{} significantly outperforms the alternative classifier in the novel environment. Leveraging the extracted features by introspective perception simplifies the learning task and allows \thisWork{} to achieve better generalizability given the same amount of training data. This is specifically a benefit for task-level failure prediction, where the volume of training data is limited due to the costly nature of acquiring data from examples of robot failures.

%% file: conclusion.tex
\section{Conclusion} \label{sec:conclusion}
In this paper, we introduced \thisWork{}, a framework for integrating introspective perception with path planning in order to learn to reduce robot navigation failures in the deployment environment and with limited amount of training data. We empirically demonstrated that by leveraging introspective perception \thisWork{} can learn a navigation competence predictor model that generalizes to novel environments and results in significantly reduced frequency of navigation failures. \thisWork{} currently addresses the problem of robot global path planning on a coarse navigation map of the environment. As future directions, the \thisWork{} framework can be extended to support competence-aware local motion planning as well as high-level task planning for mobile robots.

%% file: root_journal.bbl
\begin{thebibliography}{10}
\providecommand{\url}[1]{#1}
\csname url@rmstyle\endcsname
\providecommand{\newblock}{\relax}
\providecommand{\bibinfo}[2]{#2}
\providecommand\BIBentrySTDinterwordspacing{\spaceskip=0pt\relax}
\providecommand\BIBentryALTinterwordstretchfactor{4}
\providecommand\BIBentryALTinterwordspacing{\spaceskip=\fontdimen2\font plus
\BIBentryALTinterwordstretchfactor\fontdimen3\font minus
  \fontdimen4\font\relax}
\providecommand\BIBforeignlanguage[2]{{%
\expandafter\ifx\csname l@#1\endcsname\relax
\typeout{** WARNING: IEEEtran.bst: No hyphenation pattern has been}%
\typeout{** loaded for the language `#1'. Using the pattern for}%
\typeout{** the default language instead.}%
\else
\language=\csname l@#1\endcsname
\fi
#2}}

\bibitem{hawes2017strands}
N.~Hawes, C.~Burbridge, F.~Jovan, L.~Kunze, B.~Lacerda, L.~Mudrova, J.~Young,
  J.~Wyatt, D.~Hebesberger, T.~Kortner, \emph{et~al.}, ``The strands project:
  Long-term autonomy in everyday environments,'' \emph{IEEE Robotics \&
  Automation Magazine}, vol.~24, no.~3, pp. 146--156, 2017.

\bibitem{costante2016perception}
G.~Costante, C.~Forster, J.~Delmerico, P.~Valigi, and D.~Scaramuzza,
  ``Perception-aware path planning,'' \emph{arXiv preprint arXiv:1605.04151},
  2016.

\bibitem{rabiee2019ivoa}
S.~{Rabiee} and J.~{Biswas}, ``{IVOA}: Introspective vision for obstacle
  avoidance,'' in \emph{IEEE/RSJ International Conference on Intelligent Robots
  and Systems (IROS)}, 2019, pp. 1230--1235.

\bibitem{rabiee2020ivslam}
S.~Rabiee and J.~Biswas, ``{IV-SLAM}: Introspective vision for simultaneous
  localization and mapping,'' in \emph{Conference on Robot Learning (CoRL)},
  2020.

\bibitem{bajcsy1988active}
R.~Bajcsy, ``Active perception,'' \emph{Proceedings of the IEEE}, vol.~76,
  no.~8, pp. 966--1005, 1988.

\bibitem{aloimonos1988active}
J.~Aloimonos, I.~Weiss, and A.~Bandyopadhyay, ``Active vision,''
  \emph{International journal of computer vision}, vol.~1, pp. 333--356, 1988.

\bibitem{krotkov1988focusing}
E.~Krotkov, ``Focusing,'' \emph{International Journal of Computer Vision},
  vol.~1, no.~3, pp. 223--237, 1988.

\bibitem{krainin2011autonomous}
M.~Krainin, B.~Curless, and D.~Fox, ``Autonomous generation of complete 3d
  object models using next best view manipulation planning,'' in
  \emph{International Conference on Robotics and Automation (ICRA)}, 2011.

\bibitem{sadat2014feature}
S.~A. Sadat, K.~Chutskoff, D.~Jungic, J.~Wawerla, and R.~Vaughan,
  ``Feature-rich path planning for robust navigation of mavs with mono-slam,''
  in \emph{International Conference on Robotics and Automation (ICRA)}, 2014,
  pp. 3870--3875.

\bibitem{deng2018feature}
X.~Deng, Z.~Zhang, A.~Sintov, J.~Huang, and T.~Bretl, ``Feature-constrained
  active visual slam for mobile robot navigation,'' in \emph{International
  Conference on Robotics and Automation (ICRA)}, 2018.

\bibitem{jasour2019risk}
A.~M. Jasour and B.~C. Williams, ``Risk contours map for risk bounded motion
  planning under perception uncertainties.'' in \emph{Robotics: Science and
  Systems}, 2019.

\bibitem{shah2018airsim}
S.~Shah, D.~Dey, C.~Lovett, and A.~Kapoor, ``Airsim: High-fidelity visual and
  physical simulation for autonomous vehicles,'' in \emph{Field and service
  robotics}, 2018, pp. 621--635.

\bibitem{barbosa2021risk}
F.~S. Barbosa, B.~Lacerda, P.~Duckworth, J.~Tumova, and N.~Hawes, ``Risk-aware
  motion planning in partially known environments,'' in \emph{IEEE Conference
  on Decision and Control (CDC)}, 2021.

\bibitem{chung2019risk}
J.~J. Chung, A.~J. Smith, R.~Skeele, and G.~A. Hollinger, ``Risk-aware graph
  search with dynamic edge cost discovery,'' \emph{The International Journal of
  Robotics Research}, vol.~38, no. 2-3, pp. 182--195, 2019.

\bibitem{saxena2017learning}
D.~M. Saxena, V.~Kurtz, and M.~Hebert, ``Learning robust failure response for
  autonomous vision based flight,'' in \emph{International Conference on
  Robotics and Automation (ICRA)}, 2017, pp. 5824--5829.

\bibitem{guruau2018learn}
C.~Gur{\u{a}}u, D.~Rao, C.~H. Tong, and I.~Posner, ``Learn from experience:
  probabilistic prediction of perception performance to avoid failure,''
  \emph{The International Journal of Robotics Research}, vol.~37, no.~9, pp.
  981--995, 2018.

\bibitem{lacerda2014optimal}
B.~Lacerda, D.~Parker, and N.~Hawes, ``Optimal and dynamic planning for markov
  decision processes with co-safe ltl specifications,'' in \emph{IEEE/RSJ
  International Conference on Intelligent Robots and Systems (IROS)}, 2014, pp.
  1511--1516.

\bibitem{krajnik2017fremen}
T.~Krajn{\'\i}k, J.~P. Fentanes, J.~M. Santos, and T.~Duckett, ``Fremen:
  Frequency map enhancement for long-term mobile robot autonomy in changing
  environments,'' \emph{IEEE Transactions on Robotics}, vol.~33, no.~4, pp.
  964--977, 2017.

\bibitem{vintr2019spatio}
T.~Vintr, Z.~Yan, T.~Duckett, and T.~Krajn{\'\i}k, ``Spatio-temporal
  representation for long-term anticipation of human presence in service
  robotics,'' in \emph{International Conference on Robotics and Automation
  (ICRA)}, 2019, pp. 2620--2626.

\bibitem{basich2020learning}
C.~Basich, J.~Svegliato, K.~H. Wray, S.~Witwicki, J.~Biswas, and
  S.~Zilberstein, ``Learning to optimize autonomy in competence-aware
  systems,'' in \emph{Proceedings of the 19th International Conference on
  Autonomous Agents and MultiAgent Systems}, 2020, pp. 123--131.

\bibitem{zhang2014predicting}
P.~Zhang, J.~Wang, A.~Farhadi, M.~Hebert, and D.~Parikh, ``Predicting failures
  of vision systems,'' in \emph{Proceedings of the IEEE Conference on Computer
  Vision and Pattern Recognition}, 2014, pp. 3566--3573.

\bibitem{daftry2016}
S.~Daftry, S.~Zeng, J.~A. Bagnell, and M.~Hebert, ``Introspective perception:
  Learning to predict failures in vision systems,'' in \emph{IEEE/RSJ
  International Conference on Intelligent Robots and Systems (IROS)}, 2016, pp.
  1743--1750.

\bibitem{pulli2012real}
K.~Pulli, A.~Baksheev, K.~Kornyakov, and V.~Eruhimov, ``Real-time computer
  vision with opencv,'' \emph{Communications of the ACM}, vol.~55, no.~6, pp.
  61--69, 2012.

\end{thebibliography}
